\documentclass[11pt]{article}

\usepackage[preprint]{acl}
\usepackage{times}
\usepackage{latexsym}
\usepackage[T1]{fontenc}
\usepackage[utf8]{inputenc}
\usepackage{microtype}
\usepackage{graphicx}
\usepackage{booktabs}
\usepackage{amsmath}
\usepackage{amssymb}
\usepackage{enumitem}
\usepackage{xcolor}

\title{Cross-Task Dissociation in Frontier Vision-Language Model Theory of Mind}

\author{Kejia Zhang$^{*}$ \quad Youran Sun$^{*}$ \quad Chugang Yi \quad Haizhao Yang$^{\dagger}$ \\
  University of Maryland, College Park \\}

\begin{document}
\maketitle
\begingroup
\renewcommand{\thefootnote}{}
\footnotetext{$^*$Equal contribution. $^\dagger$Corresponding author. Emails: Youran Sun, \texttt{sun1245@umd.edu}; Haizhao Yang, \texttt{hzyang@umd.edu}.}
\endgroup

\begin{abstract}
Do frontier vision-language models present a coherent Theory-of-Mind (ToM) profile across tasks, matching the same human reference group, or does that profile fragment from one paradigm to the next?
We evaluate a shared panel of nine frontier VLMs on two psychology-derived benchmarks: the Keysar Director Task (visual perspective-taking under egocentric interference) and the Frith-Happé animated triangles scored with the Castelli rubric (intention attribution from pure motion).
On the Director Task, without chain-of-thought, the panel makes the egocentric error on 78\% of trials like children rather than adults; variation is substantial across models, and reasoning rescues several models.
On the triangles, the panel under-attributes intention: its ToM profile sits more than three times closer to the high-functioning-autistic-adult (HF-ASD) mean than to the typical-development-adult (TD) mean, while Goal-Directed and Random stay near TD.
No model is nearest TD on both tasks; the model that looks adult-like on the Director Task falls on the HF-ASD side on the triangles, and the most TD-like model on the triangles is child-like on the Director Task.
We report group-level descriptions, not diagnostic labels for any model.
\end{abstract}

\section{Introduction}
\label{sec:intro}

A user shows a vision-language model (VLM) a tabletop scene shared with another viewer.
The viewer sees fewer blocks than the model, so the model must track what that viewer can see before acting.
Psychology research on Theory of Mind (ToM) studies the capacity to reason about another person's perceptual or mental state.
It also provides mature reference profiles across developmental and clinical groups \citep{frith_happe_2000,castelli_2000,castelli_2002,keysar_2000}.
For VLMs, a single ToM score is not enough.
The question is whether one model jointly aligns with a single adult human reference profile across ToM tasks, or fragments task by task.

Recent VLM and large language model ToM benchmarks leave two gaps: they often rely on naturalistic stimuli whose faces, dialogue, and scene context afford social-cue shortcuts, and they probe one ToM sub-capacity at a time.
As a result, they cannot show whether a single model's ToM fragments across complementary facets (Section~\ref{sec:related}).

We address this gap by pairing the Keysar Director Task for visual perspective-taking with the Frith-Happ\'e animated triangles for abstract intention attribution \citep{keysar_2000,frith_happe_2000,castelli_2000}.
The Director Task asks whether a model can act from another viewer's visual access.
The animated-triangles task asks whether a model can infer intention from abstract motion.
We choose this pair because both tasks come from psychology, test complementary ToM sub-capacities, and minimize social-cue shortcuts.
The same frontier model panel evaluates both tasks (Section~\ref{sec:experiments}).
Section~\ref{sec:two_benchmarks} defines a joint coordinate system for comparing each model profile with published human reference profiles.

\begin{figure*}[t]
    \centering
    \includegraphics[width=0.8\textwidth]{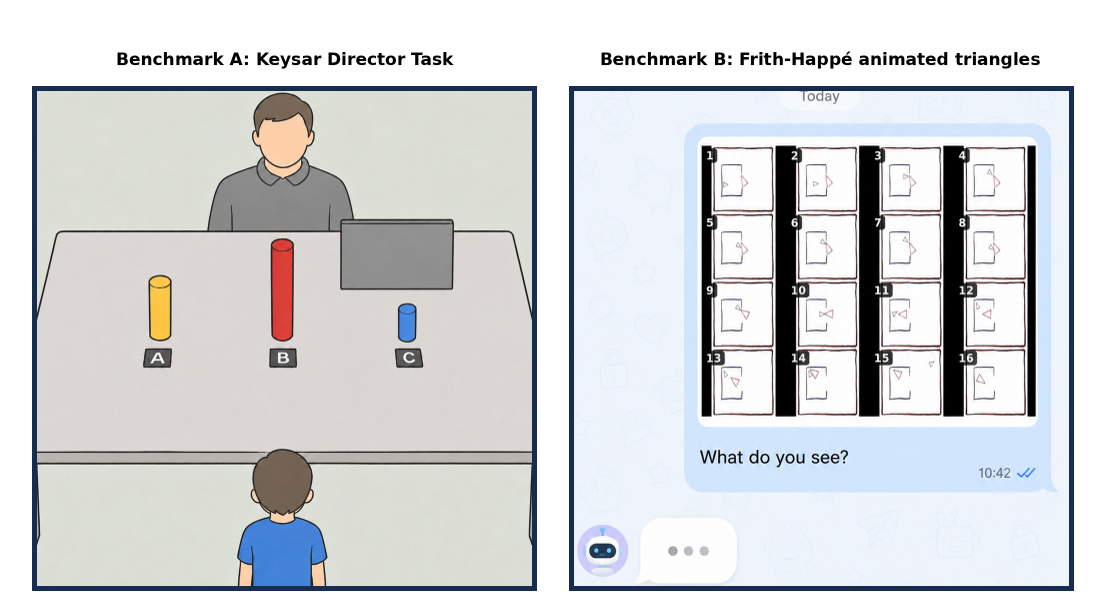}
    \caption{Schematic of the two-benchmark cross-task design. The Director Task adapts \citet{keysar_2000} and probes visual perspective-taking through the explicit-implicit gap $x = P(\text{sub-prompt a correct}) - P(\text{sub-prompt b correct})$. The animated-triangles task adapts Frith-Happ\'e clips and probes abstract intention attribution through the ToM-condition (Intent, Approp) profile. The right panel previews the joint cross-task plane (Section~\ref{sec:cross_task}).}
    \label{fig:conceptual_hero}
\end{figure*}

The analysis treats human groups as reference profiles, not diagnostic labels for models.
For each task, we compare model profiles with the relevant human anchors.
Across tasks, we report nearest-reference agreement and rank association (Section~\ref{sec:cross_task}), keeping the focus on cross-task coherence rather than pass/fail ToM claims.

Our main contributions are as follows:
\begin{itemize}[leftmargin=*,topsep=2pt,itemsep=2pt]
\item We introduce a cross-task dissociation benchmark suite for VLM ToM, with the comparison to prior benchmarks in Table~\ref{tab:prior_vs_ours}.
\item We release two reproducible psychology-grounded benchmarks: a Keysar Director Task adaptation and a Frith-Happ\'e animated-triangles adaptation \citep{keysar_2000,frith_happe_2000,castelli_2000,dureux_2023}.
\item We report a shared model-panel evaluation and cross-task reference-agreement analysis (Sections~\ref{sec:experiments} and \ref{sec:cross_task}).
\end{itemize}

\section{Related Work}
\label{sec:related}

\paragraph{ToM reference profiles in psychology.}
The Heider and Simmel film first showed that adults read intentions into animated geometric shapes \citep{heider_simmel_1944}.
Most viewers describe the shapes as agents with social goals.
Developmental psychology then built structured paradigms with age-graded and clinical-group profiles.
Examples include Three-Mountain perspective-taking \citep{piaget_1956}, Level-1/Level-2 perspective-taking \citep{flavell_1981}, the Keysar Director Task for joint action \citep{keysar_2000,apperly_2010}, and Frith-Happ\'e animated triangles with the Castelli rubric \citep{frith_happe_2000,castelli_2000,castelli_2002}.
Perspective-taking asks whether a person suppresses a privileged view when communicating with a less-informed partner.
Animated-triangles attribution asks whether a person reads goals and mental states from motion alone.
Both paradigms report group means for typically-developing adults (TD-adult), high-functioning autism-spectrum adults (HF-ASD-adult), and age-graded children cohorts \citep{castelli_2002,apperly_2010,dumontheil_2010,white_2011,livingston_2021,andersen_2022,begeer_2010,epley_2004}.
Within a paradigm, these profiles separate one human group from another.
That separation makes the published means useful anchors for frontier VLM panel profiles.
We use those anchors descriptively, not as diagnostic categories for models.

\paragraph{Existing VLM and LLM ToM benchmarks.}
VLM and large language model (LLM) ToM benchmarks differ in modality and sub-capacity, but one shared panel can show only so much under current designs.
Four naturalistic VLM benchmarks retain social cues: household scenes and bodies in MMToM-QA \citep{jin_mmtomqa_2024}, egocentric body cues in EgoToM \citep{meta_egotom_2025}, character faces and dialogue in MoMentS \citep{moments_2025}, and indoor affordances in MINDCUBE \citep{mindcube_2025}.
These cues may offer a route to the answer without explicit mental-state inference.
They also make it harder to separate mental-state inference from social-pattern matching.
Our Director Task removes objects and bodies with labeled abstract blocks.
Our animated-triangles task removes faces and dialogue with geometric motion.
These benchmarks also isolate one sub-capacity, so they cannot test whether a model fragments across complementary facets.
\citet{gao_vlms_see_2024} take the opposite design choice with an abstract three-jar perspective-taking probe, but they test only visual perspective-taking.
We add intention attribution under the same model panel, making cross-task dissociation analysis possible in a controlled shared-panel design.
No prior VLM ToM benchmark, naturalistic or abstract, pairs two complementary psychology-derived sub-tasks under a shared model panel.

\paragraph{Cross-task dissociation as the scientific posture.}
Human ToM is not a single capacity.
It includes perspective-taking, intention attribution, false-belief reasoning, affective mentalising, and second-order belief.
These sub-capacities can dissociate across populations and development.
The reference profiles above provide such within-paradigm signatures.
On animated triangles, TD-adult and HF-ASD-adult differ on ToM-condition Intentionality (Intent) but coincide on Goal-Directed (GD).
On the Director Task, young children and adults differ on the explicit-implicit gap but converge when perspective representation is explicitly cued.
A single-task VLM benchmark cannot detect analogous within-model dissociation, because the comparison requires two paradigms under one panel.
We therefore evaluate one perspective-taking paradigm and one intention-attribution paradigm on the same nine-model frontier panel.
Cross-task analysis is the headline outcome rather than a follow-up ablation.
Dissociation is the target phenomenon, not a secondary error analysis.

\paragraph{Positioning of our contribution.}
We reduce social-cue reliance per benchmark with block-and-color stimuli on the Director Task and abstract geometric trajectories on the animated-triangles task.
We discuss two residual shortcut risks in Appendix~\ref{sec:appendix_shortcuts}: motion-pattern matching on animated triangles and color-position matching on the Director Task.
Table~\ref{tab:prior_vs_ours} compares prior VLM ToM benchmarks with ours along four axes.
The table makes the contrast explicit rather than leaving it to prose.

\begin{table*}[t]
\centering
\footnotesize
\setlength{\tabcolsep}{4pt}
\begin{tabular}{p{1.8cm}p{3.6cm}p{2.8cm}p{3.5cm}p{2.6cm}}
\toprule
Benchmark & Sub-capacity measured & Stimulus type & Social-cue confounds reduced & Paired-task cross-task design \\
\midrule
MMToM-QA & belief and goal & naturalistic video + text & no (face, dialogue) & no \\
EgoToM & belief and future action & egocentric video & no (real scene context) & no \\
MoMentS & multiple ToM types & narrative film clips & no (face, dialogue, music) & no \\
MINDCUBE & cognitive map and perspective & 3D rendered & partial (abstract) & no \\
\citet{gao_vlms_see_2024} & visual perspective-taking only & abstract three-jar scenes & yes (abstract) & no (one task) \\
\midrule
\textbf{Ours} & \textbf{perspective-taking + intention attribution} & \textbf{tabletop blocks + animated triangles} & \textbf{yes (color and motion shortcuts noted)} & \textbf{yes (paired tasks)} \\
\bottomrule
\end{tabular}
\caption{Prior VLM Theory-of-Mind benchmarks vs.\ ours along sub-capacity, stimulus type, residual social-cue reduction, and paired-task design. Prior benchmarks are discussed and cited in Sec.~\ref{sec:related}; our residual shortcuts (color-position, motion-pattern) are detailed in Appendix~\ref{sec:appendix_shortcuts}.}
\label{tab:prior_vs_ours}
\end{table*}

\section{Two Benchmarks}
\label{sec:two_benchmarks}

Section~\ref{sec:benchmarkA} defines the Director Task and its split between explicit perspective representation and the implicit know-but-don't-use trap.
The animated-triangles task and its 2D Intent/Appropriateness (Approp) profile follow in Section~\ref{sec:benchmarkB}.
The joint coordinate system appears in Section~\ref{sec:joint_coords}; Section~\ref{sec:experiments} reports panel and calibration details.

\subsection{The Keysar Director Task for Visual Perspective-Taking under Action}
\label{sec:benchmarkA}

The Director Task replaces natural social scenes with block-and-color tabletop stimuli, directly reducing face, dialogue, and scene-context cues.
Its four scored outcomes split explicit perceptual perspective representation from the implicit know-but-don't-use trap.
This split lets one task report a sub-capacity profile rather than a single aggregate score.

Stimuli are programmatically generated three-dimensional tabletop scenes with colored blocks and a director figure on the far side.
A vertical opaque partition occludes one block from the director while leaving it visible to the tested model, creating a Keysar-style privileged-information asymmetry \citep{keysar_2000}.
Each block carries a ground letter label (A, B, C) and a distinct color.
The letter-and-color grounding lets the tested model refer to a block by either cue.

In one representative scene, the tested model sees small, medium, and large blocks.
The occluder hides the large block from the director, who sees only the small and medium blocks.
The utterance ``move the largest block to the right'' therefore identifies the medium block from the director's perspective.
A tested model that uses its own view picks the occluded block, the egocentric error.

For each scene the tested model receives three independent prompts with color-word answers.
Prompt (a) asks which blocks the director can see, testing perceptual perspective representation.
Prompt (b) is a single-select that asks which block to move under a director utterance, such as ``the small block''.
The utterance is ambiguous if the tested model considers all visible blocks.
It becomes unique once the tested model adopts the director's perspective.
Prompt (b) is the know-but-don't-use trap and implicit ToM probe.
Prompt (c) is a two-step explicit gate: c.Q1 re-asks (a), and c.Q2 asks which block to move.
The per-scene outcome is a four-tuple of binary correctness $(a, b, c.\text{Q1}, c.\text{Q2})$.
We also mark egocentric errors on (b) and c.Q2.

The benchmark reports per-sub-prompt accuracy and the explicit-implicit gap $P(\text{a correct}) - P(\text{b correct})$ \citep{gu_simpletom_2026}.
The evaluation harness and temporal scheduling discipline appear in Appendix~\ref{sec:appendix_firewall}.

\subsection{The Frith-Happé Animated Triangles for Abstract Intention Attribution}
\label{sec:benchmarkB}

The animated-triangles task keeps the social surface minimal: silent abstract motion, no faces, no speech, and no social labels.
Composite stills give every tested model the same temporal evidence; the judge firewall separates generation from scoring.
We retain the Castelli rubric so we can compare tested-model profiles with published human reference groups.

Stimuli are the Frith-Happ\'e animated-triangles clips re-edited by \citet{dureux_2023} under CC BY 4.0.
The clip family contains balanced ToM, GD, and Random conditions.
In ToM clips, one triangle persuades, mocks, or deceives another.
In GD clips, one triangle chases or follows another.
In Random clips, the triangles drift independently.
Each clip is approximately twenty seconds of silent abstract motion.

One representative ToM clip shows two triangles in a rectangular enclosure.
One shape persistently follows the other, while the other bobs and changes direction in response.
Human raters often describe this sequence with mentalising words such as ``coaxing'' or ``mocking'' rather than literal kinematic descriptions.

Each clip is presented to the tested model as one composite still image.
The harness samples frames uniformly and tiles them row-major.
Section~\ref{sec:experiments} reports the frame count and grid layout used in the main run.
Appendix~\ref{sec:appendix_frame_selection} reports the frame-representation calibration.
The tested model describes in free text what happens in the animation.
The prompt contains no condition label and no ToM vocabulary.

Scoring follows the \citet{castelli_2000} Appendix-2 rubric, with Intent (0--5) and Approp (0--3) dimensions.
A separate LLM judge panel scores each description; Appendix~\ref{sec:appendix_firewall} describes the rubric and code-enforced harness.
The harness hides the tested model identity from the judge and hides the rater directory from the tested model.
Appendices~\ref{sec:appendix_rubric_validation} and \ref{sec:appendix_judge_ablation} report rubric validation and judge-panel checks.
The raw scoring outputs flag same-family (judge, tested model) pairs.
Appendix~\ref{sec:appendix_judging} reports the same-family judge-control analysis.
The judge sees the clip's condition label and script semantics, following Castelli's human-rater protocol, but not the tested model identity.
The judge prompt uses the anchor-blinded Castelli rubric without any per-clip Castelli-mean overlay (Appendix~\ref{sec:appendix_judging}).

\paragraph{Aggregation Pipeline.}
For each (tested model, clip, judge) cell, we obtain an (Intent, Approp) score.
For each tested model $M_i$, we take the judge median and then the mean over ToM clips.
This yields a 2D ToM-condition profile:
\[
\operatorname{profile}_T(M_i)
= (\operatorname{Intent}_\text{ToM}, \operatorname{Approp}_\text{ToM}) .
\]
\citet{castelli_2002} define the analogous human profile $\operatorname{profile}_T(H_j)$ for each reference group.
The profile distance to human group $H_j$ is:
\[
d_T(M_i,H_j)
= \left\| \operatorname{profile}_T(M_i) - \operatorname{profile}_T(H_j) \right\|_2 .
\]
Distances use 2D ToM-condition (Intent, Approp) space.
This metric feeds the per-benchmark analysis in Section~\ref{sec:experiments} and the joint analysis in Section~\ref{sec:cross_task}.
Per-condition GD and Random profiles appear in Appendix~\ref{sec:appendix_percondition}.

\subsection{Joint Coordinate System and Cross-Task Analysis}
\label{sec:joint_coords}

The joint coordinate system makes the cross-task comparison explicit.
Shared coordinates let us visualize dissociation, measure nearest-reference switches, and report per-model disagreement.

Each model or human reference group receives a joint coordinate $(x, y)$ for Figure~\ref{fig:joint_dissociation}.
Human means come from \citet{castelli_2002} for animated triangles and from \citet{begeer_2010} and \citet{dumontheil_2010} for the Director Task.
The Director-Task coordinate $x$ is the explicit-implicit gap $P(\text{a correct}) - P(\text{b correct})$.
The same scalar definition applies to every model and human reference group.
For groups with no published explicit-implicit gap, we reconstruct it from published a-question and b-question accuracies on the closest variant.
Appendix~\ref{sec:appendix_fallback} lists the per-group sources.
Groups for which neither value is computable are excluded from Table~\ref{tab:nearest_ref_a2}.
They are also excluded from the Figure~\ref{fig:joint_dissociation} X axis.

The animated-triangles coordinate $y$ is the Figure~\ref{fig:joint_dissociation} Y axis only.
It is the ToM-condition profile distance to TD-adult.
The Director-Task nearest-reference rule is $\operatorname{nearest\_ref}(M_i, D) = \arg\min_j | x(M_i) - x(H_j) |$.
The animated-triangles nearest-reference rule uses the 2D profile distance:
\[
\operatorname{nearest\_ref}(M_i, T)
= \arg\min_j d_T(M_i,H_j).
\]
For each model we report the pair $(\operatorname{nearest\_ref}(M_i, D), \operatorname{nearest\_ref}(M_i, T))$.
Table~\ref{tab:nearest_ref_a2} also reports a binary disagreement flag.
The analysis is descriptive rather than inferential.

\section{Experiments}
\label{sec:experiments}

\subsection{Experimental Setup}
\label{sec:setup}

\paragraph{Model panel.}
The tested-model panel contains nine frontier VLMs from five labs, all accessed through a unified gateway.
The models are claude-opus-4.7, claude-sonnet-4.6, gpt-5.4, gpt-5.5, gemini-3.0-pro, gemini-3.5-flash, grok-4.3, kimi-k2.6, and qwen-3.5-plus.

\paragraph{Director Task.}
The Director-Task block contains twelve director-perspective scenes.
Each scene yields three API calls (sub-prompts a, b, c); c yields two scored outcomes (c.Q1 and c.Q2).
The scored outcomes per model per seed are $12 \times (1 + 1 + 2) = 48$.

\paragraph{Animated triangles.}
The animated-triangles stimulus set contains twelve Frith-Happ\'e clips, four per condition (ToM, GD, Random).
Each clip appears as a composite image with sixteen uniformly sampled frames in a $4 \times 4$ row-major grid.
Each cell carries a $1$--$16$ temporal-order badge; Appendix~\ref{sec:appendix_ablation} reports the badge-free variant.
Appendix~\ref{sec:appendix_frame_selection} reports the calibration for frame count, grid layout, and per-cell resolution.

\paragraph{Judge panel.}
Animated-triangles free-text responses are scored by a three-judge cross-vendor LLM panel: claude-haiku-4.5, gemini-2.5-flash, and qwen2.5-72b-instruct.
Each judge runs at temperature zero with no chain-of-thought (CoT).
Section~\ref{sec:benchmarkB} defines the harness protocol.
The panel was selected with a fourteen-anchor rubric validation battery and a seven-candidate judge ablation (Appendices~\ref{sec:appendix_rubric_validation} and \ref{sec:appendix_judge_ablation}).
An earlier judge prompt included a per-clip Castelli-mean overlay.
The audit found that overlay made the VLM-versus-Castelli comparison partially circular.
Production scoring strips the overlay while retaining the rubric and condition-label/script-semantics disclosure (Appendix~\ref{sec:appendix_judging}).

\paragraph{Replication.}
Both benchmarks use three independently seeded tester trials per (tester, item) cell ($K_\text{tester} = 3$).
We report the mean across trials and, for animated triangles, across judges, rounded to each cell's native integer scale.
For binary Director-Task cells, the rounded mean equals majority vote across trials.
Appendix Table~\ref{tab:benchmarkA_per_model} therefore reports integer counts out of $12$, not thirds-resolution counts out of $36$.
Each replication has $9 \times 12 \times 4 = 432$ Director-Task outcomes and $9 \times 12 \times 3 = 324$ animated-triangles judge cells.
Both benchmarks use three replications.
To decorrelate API calls from time-of-day server-load variance, batches are scheduled across non-contiguous time windows (Appendix~\ref{sec:appendix_firewall}).

\subsection{Director Task Results}
\label{sec:benchmarkA_results}

\begin{figure*}[t]
    \centering
    \includegraphics[width=0.8\textwidth]{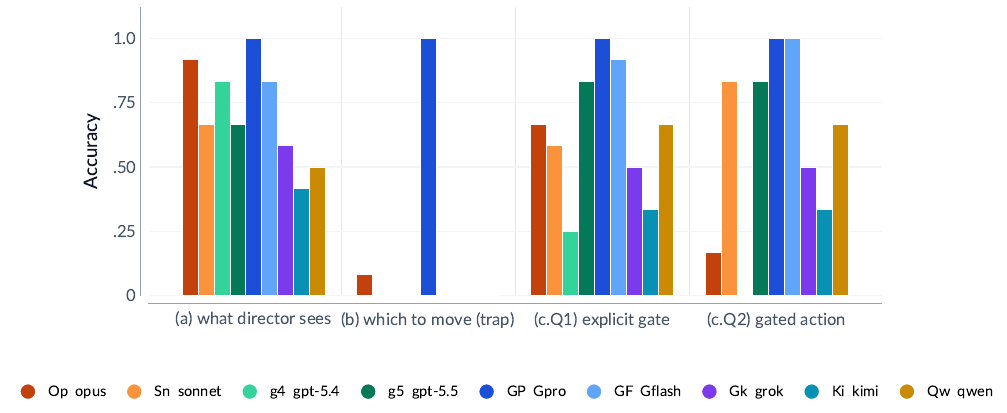}
    \caption{Director-Task per-model accuracy on the four color-word sub-prompts (Sec.~\ref{sec:benchmarkA}). The panel contains nine models, twelve canonical scenes, and three trials per (model, scene, sub-prompt) cell. The (a)--(b) gap is the explicit-implicit gap. Seven of nine models score zero on (b), claude-opus-4.7 scores $1/12$, and gemini-3.0-pro is the exception at $12/12$. Model colors match Figs.~\ref{fig:benchmarkB_tom_profile}, \ref{fig:joint_dissociation}, and \ref{fig:per_condition_panel}. Per-model accuracies appear in Appendix Table~\ref{tab:benchmarkA_per_model}.}
    \label{fig:benchmarkA_accuracy}
\end{figure*}

On the canonical Director-Task block (Figure~\ref{fig:benchmarkA_accuracy}), the panel scores $71.3\%$ on the explicit visibility multi-select (a) but collapses to $12.0\%$ on the implicit action prompt (b), an explicit-implicit gap of $59.3$ percentage points.
The collapse matches the egocentric error in the human Director-Task literature \citep{keysar_2000,apperly_2010,dumontheil_2010}: on $77.8\%$ of (b)-cells the model moves the privileged-view block.
Seven of nine models score $0/12$ on (b), and claude-opus-4.7 scores $1/12$; gemini-3.0-pro is the exception at $12/12$.

The explicit gate (c) only partly reopens the trap.
Re-asking visibility (c.Q1) lifts the panel to $63.9\%$ and the gated action (c.Q2) to $59.3\%$, but the per-model recovery $P(\text{c.Q2}) - P(b)$ splits sharply: three models recover by $\geq 80$ points (claude-sonnet-4.6 $0/12 \to 10/12$, gemini-3.5-flash $0/12 \to 12/12$, gpt-5.5 $0/12 \to 10/12$) and qwen-3.5-plus by $67$ ($0/12 \to 8/12$), while gpt-5.4 stays at $0/12$.
Perspective-use thus unblocks under the gate for some models but not others.

Against published human references, the panel-mean gap of $0.59$ falls between the TD-adult mean of $\approx 0.44$ (\citet{begeer_2010} controls $0.43$, \citet{dumontheil_2010} adults $0.44$) and the \citet{dumontheil_2010} child range, $0.59$ (ages $14.0$--$17.7$) to $0.72$ (ages $7.3$--$9.7$); the HF-ASD-adult gap is smaller still at $0.34$, consistent with the rule-based heuristic reported there \citep{begeer_2010}.
A reach-based variant brackets the same child--adult separation \citep{epley_2004}: young children (n=33, mean age $6.2$\,y) make $52\%$ egocentric reaches versus $24\%$ for adults.
We report this correspondence descriptively, without per-model developmental-age estimates.

Two supplementary controls confirm this attribution rather than a perceptual or instruction-following floor.
A floor control (Appendix~\ref{sec:appendix_director_floor}), which makes the (b) target the addressee's own visual extremum, lifts all seven non-excluded testers (qwen-3.5-plus and kimi-k2.6 excluded for latency) to letter-level $\geq 11/12$, including the six scoring $\leq 1/12$ canonically.
A three-to-four block-count control (Appendix~\ref{sec:appendix_director_blockcount}) leaves the (b) egocentric rate flat, marking trap activation as a categorical model property rather than a graded working-memory bottleneck.

\subsection{Animated Triangles Results}
\label{sec:benchmarkB_results}

\begin{figure}[t]
    \centering
    \includegraphics[width=0.8\columnwidth]{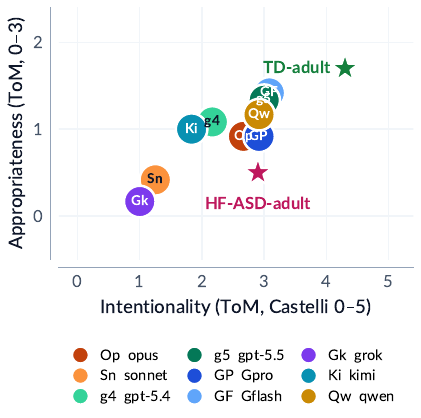}
    \caption{Per-model (Intent, Approp) profile on the animated-triangles ToM condition, using the metric from Section~\ref{sec:benchmarkB} and the anchor-blinded rubric from Appendix~\ref{sec:appendix_judging}. Each filled circle is one frontier VLM, marked by monogram and colored by lab. Stars mark the TD-adult and HF-ASD-adult group means from \citet{castelli_2002}. All nine models lie closer to HF-ASD-adult than to TD-adult on this condition.}
    \label{fig:benchmarkB_tom_profile}
\end{figure}

Figure~\ref{fig:benchmarkB_tom_profile} shows each model's ToM-condition (Intent, Approp) profile against the \citet{castelli_2002} TD-adult and HF-ASD-adult group means.

All nine models fall below both TD-adult ToM means (Intent $4.3$, Approp $1.7$). The panel-mean ToM profile $(2.31, 0.94)$ lies $2.13$ from TD-adult and $0.73$ from HF-ASD-adult $(2.9, 0.5)$; the GD mean $(1.97, 1.98)$ lies $0.51$ and $0.80$ from the two groups, and the Random mean $(0.92, 2.57)$ lies $0.88$ and $1.08$. ToM is the only condition whose panel mean sits closer to HF-ASD-adult than to TD-adult.

The asymmetry holds model by model: all nine models lie nearer HF-ASD-adult than TD-adult on ToM (per-model coordinates and per-condition breakdowns in Appendix Tables~\ref{tab:benchmarkB_distances}, \ref{tab:per_condition_distances}).
These are geometric distances to published group means, not clinical assessments.
The means are human-rated, while our scores are LLM-rated; we treat them as comparable because the judge panel recovers human-anchored exemplars within $\pm 0.5$ on the 14-anchor battery (Appendix~\ref{sec:appendix_judge_ablation}, \ref{sec:appendix_rubric_validation}).
Agreement between judges and humans on production cells is unmeasured, a noise floor we flag in Limitations.

Per-tester ranks are more heterogeneous than the panel-mean collapse implies: the strict Intent order $\text{ToM} > \text{GD} > \text{Random}$ holds for five of nine testers (claude-opus-4.7, gemini-3.0-pro, gemini-3.5-flash, gpt-5.5, qwen-3.5-plus), even though the panel-median ToM and GD Intent coincide at $2.0$.
The recoverers are the models that benefit from explicit temporal cueing; models that fail under both cued and uncued formats drive the ToM collapse (stimulus-format ablation, Appendix~\ref{sec:appendix_ablation}).

The ToM-toward-HF-ASD shift is not an artifact of the stimulus encoding: on a four-tester subset it survives both a static-keyframe and a time-reversed control (Appendix~\ref{sec:appendix_motion_probe}), with ToM nearest HF-ASD-adult in all three arms.
The time-reversed arm also reproduces the forward-time panel mean almost exactly, showing the panel is largely insensitive to temporal direction, consistent with the motion-pattern-matching shortcut of Appendix~\ref{sec:appendix_shortcuts}.

\section{Cross-Task Dissociation Analysis}
\label{sec:cross_task}

\subsection{Joint Dissociation}
\label{sec:joint_dissociation}

\begin{figure}[t]
    \centering
    \includegraphics[width=0.8\columnwidth]{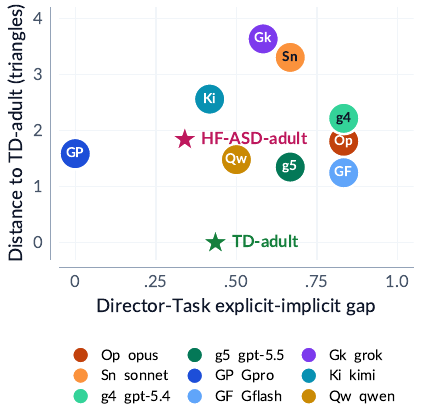}
    \caption{Joint coordinates of the nine frontier VLMs (filled circles, two-letter monograms per Fig.~\ref{fig:benchmarkB_tom_profile}) and the two adult reference groups from \citet{castelli_2002} (stars). X axis: Director-Task explicit-implicit gap $P(a)-P(b)$ (TD-adult $0.435$, HF-ASD-adult $0.340$; sources in Sec.~\ref{sec:joint_coords}). Y axis: animated-triangles ToM-condition distance to TD-adult, a visualization anchor only; the triangles nearest-reference rule uses the full 2D (Intent, Approp) space. For eight of nine models the nearest adult reference differs between the two tasks; the full cross-tabulation appears in Appendix Table~\ref{tab:nearest_ref_a2}.}
    \label{fig:joint_dissociation}
\end{figure}

Figure~\ref{fig:joint_dissociation} places the nine frontier VLMs and the relevant human reference groups in the joint coordinate system of Section~\ref{sec:joint_coords}.
The X axis is the Director-Task explicit-implicit gap; the Y axis is the animated-triangles ToM-condition profile distance to TD-adult, used only as a visualization anchor.

The per-model nearest-reference cross-tabulation is summarized below and reported in full in Appendix Table~\ref{tab:nearest_ref_a2}.
The reference set is restricted to the two adult cohorts (TD-adult and HF-ASD-adult) on both tasks, since \citet{castelli_2002} publish animated-triangles group means only for adult cohorts and no equivalent children publication exists; this is the largest reference set for which a like-for-like cross-task comparison is defined.
The wider age-graded Director-Task references of \citet{dumontheil_2010} underpin the Director-Task panel-mean gap reported above and the anchor-sensitivity analysis of Appendix~\ref{sec:appendix_anchor_sensitivity}.
Under that wider Director anchor set, most panel models reassign to a child or adolescent cohort rather than to TD-adult.
This follows from the panel-mean Director gap of $0.59$ exceeding the TD-adult anchor of $0.435$ and does not change the cross-task disagreement pattern reported below.
For eight of nine models the nearest-reference pair $(\operatorname{nearest\_ref}_A, \operatorname{nearest\_ref}_B)$ is unequal, all eight in the same direction (nearer TD-adult on the Director-Task gap scalar but nearer HF-ASD-adult on the animated-triangles 2D ToM-condition plane).
``Nearest'' here is restricted to the two adult anchors and is not a statement of absolute proximity; the panel-mean Director-Task gap of $0.59$ itself exceeds the TD-adult anchor of $0.435$.
The exception is gemini-3.0-pro, which scores $12/12$ on every Director-Task sub-prompt. Its explicit-implicit gap of $0$ sits closer to the HF-ASD-adult anchor ($0.34$) than to TD-adult ($0.435$), and it is also nearer HF-ASD-adult on the animated triangles, making it the only panel model whose nearest adult anchor is HF-ASD-adult on both tasks (a relative-distance artifact at the saturated end of the Director Task, not a substantive alignment claim).
For eight of the nine panel members no single adult reference group is jointly closest under both the Director-Task gap scalar and the animated-triangles 2D profile distance, so no single adult reference profile is nearest on both tasks for these eight models in our panel.
The animated-triangles half does not change under the stimulus-format ablation in Appendix~\ref{sec:appendix_ablation}: the plain $4 \times 4$ grid and the numbered-grid variant agree on every per-condition nearest-group verdict.

\subsection{Cross-Benchmark Rank Association}
\label{sec:rank_association}

As a secondary descriptive check, we summarize whether the per-model rankings on the two benchmarks are monotonically associated.
The Spearman rank correlation between Director-Task explicit-implicit gap and animated-triangles ToM-condition TD-distance is $\rho = -0.19$ (point estimate, $n=9$, ties broken by mid-rank).
At $n = 9$ this is an underpowered estimate (the 95\% non-parametric interval is wide enough to be uninformative), consistent with no clear monotonic rank association in this panel.

\section{Discussion and Conclusion}
\label{sec:discussion}

VLM ToM in this panel is task-dependent, so a single benchmark is not enough.
Future suites should report multi-benchmark profiles and treat profile dissociation as a main outcome.
The numbered-grid ablation supports this reading: strict Intent rank $\text{ToM} > \text{GD} > \text{Random}$ rises from zero of nine testers to five of nine without changing any per-condition nearest-group verdict (Appendix~\ref{sec:appendix_ablation}).
This points to a temporal-order bottleneck for the five recovering testers; the other four fail regardless of cue.

Three design implications follow.
Suites should pair at least two psychology-derived sub-capacities under one model panel, report per-task per-model profiles rather than single aggregate accuracy, and pair abstract with naturalistic stimuli to separate social-cue performance from explicit mental-state inference.

Frontier VLMs still fall short on these two low-social-cue ToM probes.
For eight of nine models, the nearest adult reference differs between perspective-taking and intention-attribution profile space.
The main result is therefore a panel-level dissociation, not alignment with one adult human reference profile.

\section*{Limitations}
\label{sec:limitations}
Our finding is panel-level descriptive and not an individual-model diagnostic claim.
Both benchmarks use $K_\text{tester} = 3$ independently seeded trials per (tester, item) cell, and the per-cell value reported throughout is the mean across the three trials.
Animated-triangles scoring is LLM-as-rater rather than human rater, while Castelli's reference group means come from human-rated free-text; this pipeline mismatch is a measurement noise floor that 14-anchor rubric calibration cannot fully eliminate (see Appendix~\ref{sec:appendix_judging}).
Our Director-Task implementation uses simplified block-only stimuli with color-word scoring rather than the full Director Task with physical action selection.
Two benchmarks alone do not cover the full breadth of ToM; false belief, theory-driven affect, and language-based ToM reasoning are left for future work.
The Director-Task floor and block-count controls (Appendices~\ref{sec:appendix_director_floor} and \ref{sec:appendix_director_blockcount}) cover seven of the nine testers.
qwen-3.5-plus and kimi-k2.6 were excluded from those controls for latency reasons, so the perspective-taking attribution and the categorical/graded distinction extend cleanly to the seven covered testers.
We abstain from those attributions for the excluded two.
The animated-triangles motion-probe controls (Appendix~\ref{sec:appendix_motion_probe}) cover four of the nine testers and run at $K_\text{tester}=1$ rather than the canonical $K_\text{tester}=3$; the per-arm panel-mean shifts reported there are point estimates.
We did not pre-register the analyses.
The C4 cross-task disagreement is a within-adult-anchor comparison, not a general claim that VLM behavior fails to match any human reference profile; the two-adult reference set is the largest like-for-like cross-task set defined in the published psychology literature, since \citet{castelli_2002} publish animated-triangles group means only for adult cohorts.
Appendix~\ref{sec:appendix_anchor_sensitivity} reports the asymmetric Director-side wider-anchor sensitivity, where most panel models reassign to child or adolescent anchors on the Director Task; the exact nearest-anchor labels in C4 should not be generalized beyond the available two-adult cross-task anchor set.

\section*{Ethical considerations}
\label{sec:ethics}
This paper compares frontier VLM behavior against published group-mean profiles from TD-adult, HF-ASD-adult, and age-graded children cohorts, all of which appear in the public psychology literature \citep{castelli_2002,apperly_2010,keysar_2000,begeer_2010,dumontheil_2010,epley_2004}.
We use these reference group means only as quantitative anchors for distance comparisons, and we avoid per-model diagnostic labels, developmental-age point estimates, and anthropomorphic clinical equivalences (see Section~\ref{sec:limitations}).
We use the HF-ASD-adult group mean as a reference profile in a descriptive comparison and not as a label or a value judgment on any model, system, or person.
Our benchmarks contain no human subjects data and no personally identifying information.
The Frith-Happ\'e animated-triangles clips we use are the eLife re-edits of \citet{dureux_2023}, released under CC BY 4.0.

\bibliography{custom}

\appendix
\section{Implementation Firewall Details}
\label{sec:appendix_firewall}

Both benchmarks share a code-enforced two-firewall harness that hides tester identity from judges and hides rater-side files from testers; we describe each firewall separately below rather than label the combination ``double-blind'', since the rater is deliberately given the clip's ground-truth condition label and script semantics in keeping with Castelli's original human-rater protocol.
The first firewall bars testers from reading any path under the rater directory and from reading the test-set registry that maps tester display names to backing IDs.
The second firewall bars judges from reading tester short names; only opaque tester IDs reach the judge prompt.
All run artifacts are written atomically append-only with an open-exclusive create flag so that no past run can be silently rewritten.

\paragraph{Temporal decorrelation against server-load variance.}
API call batches in this work, spanning the two main-panel benchmarks, every calibration pre-experiment, and every shuffleseed replicate, are scheduled at different times of day rather than executed in a single contiguous burst.
Time-correlated load variance on the model providers' inference servers, including throttling, queue depth, and concurrent traffic spikes, is therefore averaged across calls rather than concentrated within one window.
Concretely, the animated-triangles plain-grid and numbered-grid main panels are launched approximately 22 hours apart, the $K_\text{tester}=3$ replicates of the frame-representation pre-experiment (Appendix~\ref{sec:appendix_frame_selection}) span two calendar days, and the rubric and judge validation battery (Appendix~\ref{sec:appendix_rubric_validation}), the judge model ablation (Appendix~\ref{sec:appendix_judge_ablation}), and the Director-Task shuffleseed replicates are each launched on distinct calendar days.

The full harness, scoring scripts, and per-(tester, clip, judge) raw outputs will be released with the camera-ready.

\section{Per-Model Director Task Sub-Prompt Accuracy}
\label{sec:appendix_benchmarkA}

Table~\ref{tab:benchmarkA_per_model} reports the per-model raw correct count out of $12$ canonical Director-Task scenes on each of the four sub-prompts.
The egocentric-error rate on sub-prompt (b), reported as ``ego.\ (b)'' in the rightmost column, is the fraction of cells in which the model picks the privileged-view block that would be ambiguous for the addressee but is in fact occluded from the director.
Seven of nine models score $0$ correct on sub-prompt (b) and claude-opus-4.7 scores $1$ of $12$; the lone exception that does not exhibit the egocentric failure mode is gemini-3.0-pro at $12/12$.

\begin{table*}[t]
\centering
\small
\begin{tabular}{lccccc}
\toprule
Model & a & b & c.Q1 & c.Q2 & ego.\ (b) \\
\midrule
opus-4.7         & 11 & 1  & 8  & 2  & 9  \\
sonnet-4.6       & 8  & 0  & 7  & 10 & 9  \\
gemini-3.0-pro   & 12 & 12 & 12 & 12 & 0  \\
gemini-3.5-flash & 10 & 0  & 11 & 12 & 12 \\
gpt-5.4          & 10 & 0  & 3  & 0  & 12 \\
gpt-5.5          & 8  & 0  & 10 & 10 & 10 \\
grok-4.3         & 7  & 0  & 6  & 6  & 12 \\
kimi-k2.6        & 5  & 0  & 4  & 4  & 9  \\
qwen-3.5-plus    & 6  & 0  & 8  & 8  & 11 \\
\midrule
panel mean (\%)  & 71.3 & 12.0 & 63.9 & 59.3 & 77.8 \\
\bottomrule
\end{tabular}
\caption{Director-Task per-model raw counts (out of $12$ canonical scenes) on each sub-prompt, plus egocentric-error count on sub-prompt (b). Sub-prompt (a) is the multi-select ``which blocks does the director see''. Sub-prompt (b) is the single-select know-but-don't-use trap. Sub-prompt (c) is split into Q1 (explicit re-asking of a) and Q2 (gated action after Q1). The panel-mean Director-Task explicit-implicit gap $P(a)-P(b)$ is $59.3$ percentage points.}
\label{tab:benchmarkA_per_model}
\end{table*}

\section{Director Task Perceptual and Instruction Floor Control}
\label{sec:appendix_director_floor}

\paragraph{Design.}
We run two control arms on the canonical twelve Director-Task scenes.
The no-occluder arm removes the occluder so all three blocks are visible to both addressee and director; the (a, b, c.Q1, c.Q2) sub-prompt structure is unchanged.
The same-side-director arm keeps the occluder but moves the director to the addressee's side of the table, so the occluder does not occlude anything from the director (occluder is physically present but geometrically inert).
Both arms invert the sub-(b) ground-truth target onto the addressee's visual extremum, so a model that simply picks the visual extremum scores correctly on the control sub-(b).
The remaining failure modes are perceptual mis-identification (cannot see all three blocks, mis-counts blocks, mis-takes occluder for a block) or instruction failure (does not understand ``largest''/``smallest'').

\paragraph{Panel and pre-registered floor pass criterion.}
We run both arms on the same nine-model panel minus qwen-3.5-plus and kimi-k2.6, excluded for latency reasons (the same exclusion holds for Appendix~\ref{sec:appendix_director_blockcount}).
The seven testers retained are claude-opus-4.7, claude-sonnet-4.6, gpt-5.4, gpt-5.5, gemini-3.0-pro, gemini-3.5-flash, and grok-4.3.
The canonical-(b) attribution for the two excluded testers is therefore not covered by this control.
The floor pass criterion, fixed before running paid API calls, is per-model control sub-(b) letter-level correctness $\geq 11/12$, where letter-level marks the correct letter regardless of color naming (failure-mode $\in \{\text{correct}, \text{letter\_only\_correct}\}$).
The letter-level threshold isolates the perceptual / instruction floor from the orthogonal blue $\to$ cyan color-naming drift that appears panel-wide on the closed color vocabulary.

\paragraph{Result.}
All seven testers pass the floor on both arms (7/7 on each arm).
Table~\ref{tab:director_floor} reports per-model canonical-(b) versus control-(b) letter-level scores.

\begin{table*}[t]
\centering
\small
\begin{tabular}{lccc}
\toprule
Tester & canon-(b) letter & no-occluder (b) letter & same-side-director (b) letter \\
\midrule
claude-opus-4.7   & 1/12  & 12/12 & 12/12 \\
claude-sonnet-4.6 & 0/12  & 12/12 & 12/12 \\
gemini-3.0-pro    & 12/12 & 12/12 & 12/12 \\
gemini-3.5-flash  & 0/12  & 12/12 & 12/12 \\
gpt-5.4           & 0/12  & 12/12 & 12/12 \\
gpt-5.5           & 0/12  & 12/12 & 12/12 \\
grok-4.3          & 0/12  & 12/12 & 11/12 \\
\bottomrule
\end{tabular}
\caption{Director-Task floor control. canon-(b) is from the canonical 12-scene Director Task; the no-occluder arm removes the occluder; the same-side-director arm keeps the occluder but moves the director to the addressee side. Letter-level $=$ correct letter regardless of color name; strict pair-level results lie $2/12$ below letter-level for every cell because Q01 and Q02 expect blue and the panel reads blue as cyan. The six models with canonical $\leq 1/12$ score $\geq 11/12$ on both control arms, attributing the canonical collapse to perspective-taking rather than perceptual or instruction floor.}
\label{tab:director_floor}
\end{table*}

\paragraph{Independent side finding on the same-side-director arm.}
On the same-side-director arm, the two Claude testers show degraded sub-(a) and sub-c.Q1 identification despite passing the sub-(b) floor.
claude-opus-4.7 scores sub-(a) $7/12$ (no-occluder arm: $12/12$) and c.Q1 $4/12$; claude-sonnet-4.6 scores sub-(a) $12/12$ but c.Q1 $9/12$ and c.Q2 $7/12$.
This is consistent with the two Claude models maintaining a residual ``the occluder physically blocks the director's view'' inference even when the geometry no longer supports it.
The finding is independent of the floor result and does not affect the attribution conclusion; the floor judgment lives in sub-(b) only.

\section{Director Task Variable Block Count}
\label{sec:appendix_director_blockcount}

\paragraph{Design.}
We add a four-block variant of the Director Task structure to test whether the canonical sub-(b) failure rate scales with the number of competing blocks (graded, working-memory-style) or is categorical (binary, the model either represents the director's view or it does not).
Four blocks of distinct sizes are placed in front of the director.
Exactly one block (the global extremum under the director's instruction) is occluded from the director by the partition; the director then issues an extremum instruction (``move the largest/smallest block to the right'') and the ground-truth target is the extremum among the director-visible set (the second-largest or second-smallest of the four).
The (a, b, c.Q1, c.Q2) sub-prompt structure and the deterministic regex-based scoring are unchanged.
We sample twelve balanced four-block scenes (six largest-instruction, six smallest-instruction, with the egocentric-trap position balanced across the four block positions).
The n=3 baseline used for the matched-pair comparison is the same seven testers on the same closed-color-vocabulary scorer as the four-block run.

\paragraph{Panel.}
Same seven testers as Appendix~\ref{sec:appendix_director_floor} (qwen-3.5-plus and kimi-k2.6 excluded for latency).

\paragraph{Result.}
Table~\ref{tab:director_blockcount} reports per-tester sub-(b) direct-action correct and egocentric error rate at n=3 versus n=4 competing blocks.

\begin{table*}[t]
\centering
\small
\begin{tabular}{lcccc}
\toprule
Tester & n=3 (b) correct & n=3 (b) ego & n=4 (b) correct & n=4 (b) ego \\
\midrule
gemini-3.0-pro    & 12/12 & 0/12  & 12/12 & 0/12 \\
gpt-5.5           & 0/12  & 10/12 & 0/12  & 10/12 \\
claude-sonnet-4.6 & 0/12  & 10/12 & 0/12  & 10/12 \\
claude-opus-4.7   & 0/12  & 9/12  & 0/12  & 10/12 \\
gpt-5.4           & 0/12  & 10/12 & 0/12  & 10/12 \\
gemini-3.5-flash  & 0/12  & 10/12 & 0/12  & 10/12 \\
grok-4.3          & 0/12  & 10/12 & 3/12  & 7/12 \\
\bottomrule
\end{tabular}
\caption{Director-Task sub-(b) direct-action correct and egocentric error rate at n=3 vs n=4 competing blocks, seven-tester panel. The sole non-collapsing model at n=3 (gemini-3.0-pro) stays at $12/12$ correct under n=4. The six collapsing models stay at $0/12$ direct-action correct with $\geq 7/12$ egocentric error under n=4. Both ends of the panel are flat with respect to block count.}
\label{tab:director_blockcount}
\end{table*}

\paragraph{Take-away.}
The sub-(b) failure rate is flat with block count in this panel, neither the capable model nor the six collapsed models move with the additional competing block.
This is consistent with sub-(b) failure being a categorical model property (the model either represents the director's view in action or it does not) rather than a graded working-memory or attentional bottleneck that should grow with competing items.
The explicit-gate CoT recovery on c.Q2 does decline with the extra block for some models (claude-sonnet-4.6 $10 \to 5$, gpt-5.5 $10 \to 9$, claude-opus-4.7 $2 \to 1$), while remaining steady for gemini-3.5-flash ($12 \to 12$) and gpt-5.4 ($0 \to 0$); the recovery channel is graded for the models whose recovery is CoT-mediated, even when the underlying trap activation is categorical.

\section{Per-Condition Breakdown for the Animated Triangles}
\label{sec:appendix_percondition}

\begin{figure*}[t]
    \centering
    \includegraphics[width=\textwidth]{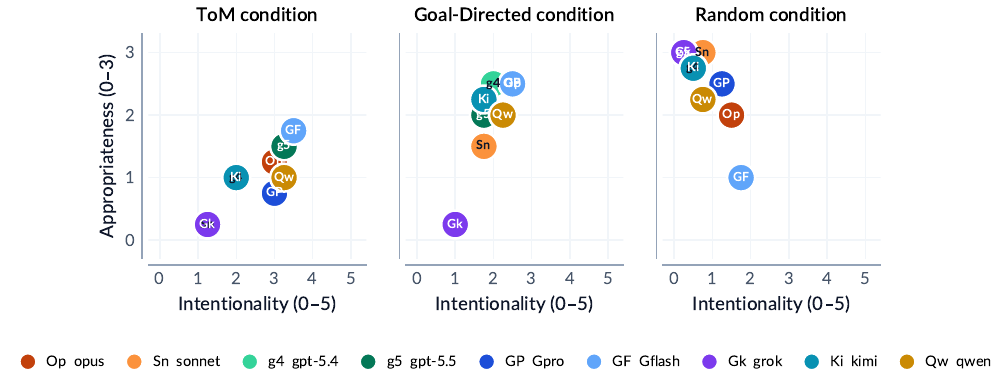}
    \caption{Per-model (Intent, Approp) profile on the animated triangles broken down by condition. Each panel shows the per-model points, colored by lab. The canonical ToM-condition panel is also reported as Figure~\ref{fig:benchmarkB_tom_profile} in the main text.}
    \label{fig:per_condition_panel}
\end{figure*}

The main text reports the animated-triangles ToM-condition (Intent, Approp) profile and the corresponding profile distances under the numbered-grid canonical metric.
Figure~\ref{fig:per_condition_panel} and Table~\ref{tab:per_condition_distances} report the per-condition (GD, Random) panels and the per-condition profile distances under the same v2 canonical metric.

\begin{table*}[t]
\centering
\small
\begin{tabular}{lcccccc}
\toprule
 & \multicolumn{2}{c}{ToM cond.} & \multicolumn{2}{c}{GD cond.} & \multicolumn{2}{c}{Random cond.} \\
\cmidrule(lr){2-3} \cmidrule(lr){4-5} \cmidrule(lr){6-7}
Model & to TD & to ASD & to TD & to ASD & to TD & to ASD \\
\midrule
opus-4.7         & 1.811 & 0.478 & 0.765 & 1.148 & 1.250 & 1.146 \\
sonnet-4.6       & 3.309 & 1.652 & 1.046 & 0.910 & 1.203 & 1.516 \\
gpt-5.4          & 2.221 & 0.937 & 0.672 & 0.992 & 1.230 & 1.383 \\
gpt-5.5          & 1.351 & 0.839 & 0.827 & 0.975 & 1.203 & 1.516 \\
gemini-3.0-pro   & 1.590 & 0.417 & 0.754 & 1.141 & 0.887 & 1.090 \\
gemini-3.5-flash & 1.249 & 0.935 & 0.843 & 1.229 & 1.347 & 0.929 \\
grok-4.3         & 3.639 & 1.929 & 1.647 & 1.476 & 1.203 & 1.548 \\
kimi-k2.6        & 2.564 & 1.178 & 0.864 & 1.217 & 1.037 & 1.351 \\
qwen-3.5-plus    & 1.483 & 0.667 & 0.389 & 0.786 & 0.902 & 1.168 \\
\bottomrule
\end{tabular}
\caption{Per-model 2D Euclidean profile distance to TD-adult and HF-ASD-adult group means under each animated-triangles condition (ToM, GD, Random), numbered-grid stimulus, anchor-blinded canonical rubric per Appendix~\ref{sec:appendix_judging}. The ToM-condition columns are the canonical animated-triangles metric (Sec.~\ref{sec:benchmarkB}). On the ToM condition, all nine models have larger distance to TD-adult than to HF-ASD-adult.}
\label{tab:per_condition_distances}
\end{table*}

\section{Animated-Triangles Motion-Probe Controls}
\label{sec:appendix_motion_probe}

\paragraph{Design.}
We run two stimulus-format controls on the canonical twelve animated-triangles clips.
The static-keyframe arm replaces each $4 \times 4$ composite with a single mid-clip frame in native $1280 \times 720$ resolution.
The time-reversed arm keeps the same sixteen-cell $4 \times 4$ composite and the $1$--$16$ temporal-order badges, but reverses the underlying frame order so that the cell labelled ``frame 1'' shows what was originally the last sampled frame; the tested model is not told the order is reversed.
Both controls use the same Castelli rubric, the same three-judge cross-vendor panel, and the same anchor-blinded canonical scoring as the main animated-triangles run.

\paragraph{Panel and replication.}
Four testers, the per-lab best animated-triangles performer for the four Western labs (claude-opus-4.7, gpt-5.5, gemini-3.5-flash, grok-4.3); qwen-3.5-plus and kimi-k2.6 are excluded for latency, the same exclusion as in Appendices~\ref{sec:appendix_director_floor} and \ref{sec:appendix_director_blockcount}.
Both controls run with $K_\text{tester} = 1$ per (tester, clip) cell; the canonical forward-time reference used here is the nine-model main run sub-selected to the same four testers.

\paragraph{Result.}
Table~\ref{tab:motion_probe} reports the per-condition panel-mean (Intent, Approp) for the three arms and the Euclidean distance to each adult reference group.

\begin{table*}[t]
\centering
\small
\begin{tabular}{llrrrl}
\toprule
Condition & Arm & (Intent, Approp) & dist.\ TD & dist.\ HF-ASD & Nearest \\
\midrule
ToM    & canonical fwd   & (2.65, 1.08) & 1.77 & 0.64 & HF-ASD \\
ToM    & time-reversed   & (2.56, 1.06) & 1.85 & 0.66 & HF-ASD \\
ToM    & static keyframe & (1.96, 0.73) & 2.53 & 0.97 & HF-ASD \\
\midrule
GD     & canonical fwd   & (1.96, 1.83) & 0.46 & 0.69 & TD     \\
GD     & time-reversed   & (1.96, 1.67) & 0.44 & 0.57 & TD     \\
GD     & static keyframe & (1.98, 1.25) & 0.62 & 0.42 & HF-ASD \\
\midrule
Random & canonical fwd   & (0.98, 2.19) & 0.62 & 0.71 & TD     \\
Random & time-reversed   & (0.81, 2.35) & 0.64 & 0.85 & TD     \\
Random & static keyframe & (1.48, 1.00) & 1.26 & 0.84 & HF-ASD \\
\bottomrule
\end{tabular}
\caption{Animated-triangles motion-probe controls on the four-tester subset (claude-opus-4.7, gpt-5.5, gemini-3.5-flash, grok-4.3, the per-lab Western frontier best on the animated triangles). The canonical-forward arm is the nine-model main run sub-selected to the same four testers. The time-reversed arm reproduces the canonical forward-time arm almost exactly on all three conditions. The static-keyframe arm shifts the ToM panel mean toward lower Intent and the GD/Random panel means toward lower Approp; ToM nevertheless remains nearer HF-ASD-adult than TD-adult under all three arms.}
\label{tab:motion_probe}
\end{table*}

\paragraph{Take-away.}
On ToM, the time-reversed and forward-time panel means coincide to within $\sim 0.1$ on both axes, and both lie at distance $\sim 0.65$ to HF-ASD-adult and $\sim 1.8$ to TD-adult; reversing temporal direction does not change the asymmetric ToM-toward-HF-ASD shift.
The static-keyframe arm reduces ToM-condition panel mean Intent from $2.65$ to $1.96$ but the panel mean still lands closer to HF-ASD-adult than to TD-adult.
A complementary effect on the non-ToM conditions is that the static-keyframe arm raises the panel mean attributed Intent on Random from $0.98$ to $1.48$ and lowers attributed Approp on GD, blurring condition distinctions; the per-tester strict Intent rank $\text{ToM} > \text{GD} > \text{Random}$ holds for only one of four testers under the static keyframe versus two of four under both grid arms.
Two readings are compatible with this pattern.
First, a large fraction of the ToM-condition attribution signal is recoverable from a single representative frame, consistent with static composition and style priors (the shape rendering, the enclosure, the relative positions of the triangles) carrying much of the mental-attribution signal.
Second, on the canonical multi-frame arm the role of the motion structure is largely to suppress over-attribution to non-ToM conditions rather than to provide the ToM signal itself.
The time-reversed result is consistent with the motion-pattern-matching residual shortcut noted in Appendix~\ref{sec:appendix_shortcuts}.

\paragraph{Caveats.}
Both controls run at $K_\text{tester}=1$ while the canonical main run uses $K_\text{tester}=3$; the controls give point estimates and the per-arm panel-mean shifts should be read as descriptive.
The static-keyframe prompt informs the tested model that it is seeing a single still frame from a short silent animation; this is a design tradeoff to keep the task framing aligned with the multi-frame arms, not an unintended leak.
The time-reversed implementation reverses the order of the sixteen sampled frames rather than re-rendering the clip frame-by-frame in reverse, so the temporal-direction signal the tested model receives is the order of the badges and of the in-cell stills, which matches what the canonical forward-time arm provides.

\section{Per-Model Distances on the Animated-Triangles ToM Condition}
\label{sec:appendix_distances}

\begin{table*}[t]
\centering
\small
\begin{tabular}{lcccc}
\toprule
Model & Intent & Approp & dist.\ TD-adult & dist.\ HF-ASD-adult \\
\midrule
opus-4.7         & 2.667 & 0.917 & 1.811 & 0.478 \\
sonnet-4.6       & 1.250 & 0.417 & 3.309 & 1.652 \\
gpt-5.4          & 2.167 & 1.083 & 2.221 & 0.937 \\
gpt-5.5          & 3.000 & 1.333 & 1.351 & 0.839 \\
gemini-3.0-pro   & 2.917 & 0.917 & 1.590 & 0.417 \\
gemini-3.5-flash & 3.083 & 1.417 & 1.249 & 0.935 \\
grok-4.3         & 1.000 & 0.167 & 3.639 & 1.929 \\
kimi-k2.6        & 1.833 & 1.000 & 2.564 & 1.178 \\
qwen-3.5-plus    & 2.917 & 1.167 & 1.483 & 0.667 \\
\bottomrule
\end{tabular}
\caption{Per-model ToM-condition (Intent, Approp) coordinates and 2D Euclidean profile distance from each frontier VLM to the TD-adult $(4.3, 1.7)$ and HF-ASD-adult $(2.9, 0.5)$ published group means on the animated triangles (canonical metric per Sec.~\ref{sec:benchmarkB}, numbered-grid stimulus, anchor-blinded canonical rubric per Appendix~\ref{sec:appendix_judging}).}
\label{tab:benchmarkB_distances}
\end{table*}

\section{Reference-Anchor Sensitivity and Fallback Sources}
\label{sec:appendix_anchor_sensitivity}
\label{sec:appendix_fallback}

\paragraph{Animated-triangles anchor.}
The animated-triangles coordinate $y$ in the joint plot is anchored to TD-adult for visualization, but the per-model nearest-reference rule (Section~\ref{sec:joint_coords}) is computed in the full 2D (Intent, Approp) plane and is therefore independent of which adult group is used as the visualization anchor.
\citet{castelli_2002} do not publish age-graded children animated-triangles group means, so we cannot extend the reference set on this task; this asymmetry between the two tasks motivates the adult-only restriction on the cross-task nearest-reference cross-tabulation in Section~\ref{sec:cross_task}.

\begin{table*}[t]
\centering
\small
\begin{tabular}{lcccc}
\toprule
Model & $x_A$ & $\operatorname{nearest\_ref}_A$ & $\operatorname{nearest\_ref}_B$ & disagree \\
\midrule
opus-4.7         & 0.833 & TD-adult     & HF-ASD-adult & yes \\
sonnet-4.6       & 0.667 & TD-adult     & HF-ASD-adult & yes \\
gpt-5.4          & 0.833 & TD-adult     & HF-ASD-adult & yes \\
gpt-5.5          & 0.667 & TD-adult     & HF-ASD-adult & yes \\
gemini-3.0-pro   & 0.000 & HF-ASD-adult & HF-ASD-adult & no  \\
gemini-3.5-flash & 0.833 & TD-adult     & HF-ASD-adult & yes \\
grok-4.3         & 0.583 & TD-adult     & HF-ASD-adult & yes \\
kimi-k2.6        & 0.417 & TD-adult     & HF-ASD-adult & yes \\
qwen-3.5-plus    & 0.500 & TD-adult     & HF-ASD-adult & yes \\
\midrule
\multicolumn{4}{r}{disagreement count} & \textbf{8 of 9} \\
\bottomrule
\end{tabular}
\caption{Per-model nearest-reference cross-tabulation (anchor-blinded canonical rubric). $x_A$: Director-Task gap $P(a)-P(b)$; $\operatorname{nearest\_ref}_A$ = closer of TD-adult ($0.435$) and HF-ASD-adult ($0.340$) under $|x - x_H|$; $\operatorname{nearest\_ref}_B$ = closer reference on the animated-triangles (Intent, Approp) ToM plane (Sec.~\ref{sec:joint_coords}). Eight of nine disagree, all nearer TD-adult on the Director Task but HF-ASD-adult on the triangles. The exception, gemini-3.0-pro, saturates the Director Task ($12/12$, gap $0$): a relative-distance artifact ($|0-0.435|>|0-0.340|$), not an alignment claim.}
\label{tab:nearest_ref_a2}
\end{table*}

\paragraph{Director-Task anchor and fallback.}
The Director-Task coordinate $x$ uses the published explicit-implicit gap per group with canonical sign $x = P(\text{a correct}) - P(\text{b correct})$, applied identically to models and human groups.
For a human group with no published explicit-implicit gap on the same paradigm, the fallback reconstructs the same scalar as published a-question accuracy minus published b-question accuracy on the closest published variant (canonical sign preserved).
The per-group sources currently in use are \citet{begeer_2010} (TD-adult controls and HF-ASD-adult), \citet{dumontheil_2010} (TD-adult and four age-graded cohorts spanning $7.3$--$17.7$ years), and \citet{epley_2004} as a qualitative bracket (4--12-year sample, reach-rate metric not directly comparable to the gap scalar and therefore plotted separately rather than as an anchor).

\paragraph{Sensitivity of $\operatorname{nearest\_ref}_A$ to the reference set.}
Under the two-adult reference set $\{\text{TD-adult}, \text{HF-ASD-adult}\}$ used in Table~\ref{tab:nearest_ref_a2}, eight of nine models map to TD-adult on the Director Task and one (gemini-3.0-pro, gap $0$) maps to HF-ASD-adult.
Under the wider reference set that additionally includes the four \citet{dumontheil_2010} age-graded cohorts (gaps $0.59$, $0.67$, $0.68$, $0.72$), six of nine models reassign to a children/adolescent cohort (the cohort at gap $0.59$ for grok-4.3 at $0.583$; the cohorts at $0.67$--$0.68$ for sonnet-4.6 and gpt-5.5 at $0.667$; the cohort at $0.72$ for opus-4.7, gemini-3.5-flash, and gpt-5.4 at $0.833$), kimi-k2.6 and qwen-3.5-plus stay at TD-adult, and gemini-3.0-pro stays at HF-ASD-adult.
Crucially, the cross-task disagreement count is robust to this anchor expansion.
The cross-task disagreement remains eight of nine because the animated-triangles nearest reference is HF-ASD-adult for all eight non-gemini-3.0-pro models, and only gemini-3.0-pro keeps the same nearest adult anchor on both tasks under either reference set.

\paragraph{Sensitivity of $\operatorname{nearest\_ref}_B$ to the Y-axis anchor.}
Swapping the Y-axis visualization anchor from TD-adult to HF-ASD-adult shifts every Y coordinate by the constant adult-adult distance ($1.844$ in 2D) but does not change the 2D Euclidean nearest-reference assignment, so $\operatorname{nearest\_ref}_B$ is invariant under this swap.

\section{Residual Shortcut Risks}
\label{sec:appendix_shortcuts}

The Director Task reduces social-cue reliance with block-and-color stimuli and an explicit letter-and-color double-grounding scheme.
A natural worry is color-position matching, in which a model uses block color or scene position alone to predict the intended target rather than reasoning about the director's perspective.
The canonical 12-scene set explicitly controls for this by rotating the color palette across three independent triples (yellow / red / blue; purple / green / orange; red / cyan / green), rotating the letter-to-color mapping per scene (A, B, C bind to different colors in different scenes), rotating which letter position carries the target block, and rotating the instruction polarity (smallest / largest).
A color-position shortcut would have to survive all four rotations to produce the per-model failure patterns we observe in Table~\ref{tab:benchmarkA_per_model}.
The animated triangles reduce social-cue reliance with abstract geometric trajectories.
A residual shortcut is motion-pattern matching, in which a model classifies trajectories by their kinematic signature alone without invoking mental-state inference; this shortcut is the target of the queued reversed-time playback ablation and is bounded but not fully eliminated by the current design.

\section{LLM-as-Rater Considerations and Blinding-Robustness Audit}
\label{sec:appendix_judging}

\paragraph{Mitigations.}
Castelli's published reference group means come from human-rated free-text; we use an LLM-as-rater jury on the same rubric.
The jury uses three different model families (Anthropic, Google, Qwen) so that no single model family dominates scoring; the cross-family choice was empirically validated in Appendix~\ref{sec:appendix_judge_ablation}.
Every rater operates under the Section~\ref{sec:benchmarkB} two-firewall protocol so no rater knows which tester produced which output (although the rater does see the clip's ground-truth condition label and script semantics, following Castelli's original protocol), and all rater outputs are written append-only so post-hoc tampering of scores is detectable.
The three production judges were further validated against the 14 human-anchored paper exemplars of Castelli (2000, 2002) under the Phase-1 battery in Appendix~\ref{sec:appendix_rubric_validation}, each achieving 100\% anchor recovery with per-anchor standard deviation equal to zero across 5 repetitions at $T=0$.

\paragraph{Information disclosure to the judge.}
Each judge call carries, per clip, three layers of overlay information on top of the SHA-locked Castelli rubric.
Layer~(i) is the ground-truth condition label (ToM, GD, or Random) for that clip.
Layer~(ii) is the animation script semantics for that clip (a one-sentence description such as ``the large triangle coaxes the small triangle out of the box'').
Layer~(i) and layer~(ii) follow the original Castelli protocol; human raters in \citet{castelli_2002} knew the animation script in order to grade Approp against the intended interaction type.
Layer~(iii) is an \texttt{Expected score range} block giving the Castelli-2002 human group-mean as a per-clip Intent target (for example, ``Expected Intent: 4--5'' on ToM clips, derived from the TD-adult ToM Intent mean of $4.3$).

\paragraph{Audit finding.}
An audit of the production judge prompt confirmed that layer~(iii) was not part of any human-rater protocol.
Inserting the Castelli human group mean as a per-clip target anchors the judge upward and makes the Stage-C comparison (VLM profile against the same Castelli group mean) partly circular.
The direction of this bias is conservative for the headline finding, since a judge anchored toward TD-adult should still let the VLM profile drift toward TD-adult, so a finding of ``VLM panel lies below TD-adult'' under this anchored judge is biased against itself.

\paragraph{Blinded L1 re-score.}
The anchor-blinded L1 re-score uses byte-identical tester responses, the same three-judge cross-vendor panel at $T=0$, the same SHA-locked rubric, and an overlay that strips only layer~(iii); layer~(i) and layer~(ii) are retained.

\paragraph{Informed-vs-anchor-blinded comparison.}
Table~\ref{tab:judge_blinding} reports the per-arm panel-level headline metrics under the two judge variants on the same $324$ canonical cells per arm.

\begin{table*}[t]
\centering
\footnotesize
\setlength{\tabcolsep}{4pt}
\begin{tabular}{p{5.6cm}cccc}
\toprule
Metric & \shortstack{plain\\informed} & \shortstack{plain\\anchor-blinded} & \shortstack{numbered\\informed} & \shortstack{numbered\\anchor-blinded} \\
\midrule
Panel median Intent on ToM / GD / Random & $2.0\,/\,2.0\,/\,1.0$ & $2.0\,/\,2.0\,/\,1.0$ & $2.0\,/\,2.0\,/\,1.0$ & $2.0\,/\,2.0\,/\,1.0$ \\
Strict rank $\text{ToM} > \text{GD} > \text{Random}$ & $1\,/\,9$ & $0\,/\,9$ & $4\,/\,9$ & $5\,/\,9$ \\
Panel mean (Intent, Approp) on ToM & $(2.33, 0.95)$ & $(2.25, 0.95)$ & $(2.46, 0.97)$ & $(2.31, 0.94)$ \\
Distance from panel mean on ToM to TD-adult & $2.10$ & $2.18$ & $1.98$ & $2.13$ \\
Distance from panel mean on ToM to HF-ASD-adult & $0.73$ & $0.79$ & $0.64$ & $0.73$ \\
ToM nearest adult reference & HF-ASD & HF-ASD & HF-ASD & HF-ASD \\
GD nearest adult reference & TD & TD & TD & TD \\
Random nearest adult reference & TD & TD & TD & TD \\
\bottomrule
\end{tabular}
\caption{Judge information-disclosure ablation. The anchor-blinded L1 variant strips only the per-clip Castelli-mean anchor from the judge prompt; everything else, including the tester responses, judge identities, rubric SHA, and aggregation logic, is held fixed. Across both stimulus-format arms (plain $4 \times 4$ grid and numbered $4 \times 4$ grid) the nearest-group verdict per condition is unchanged, the panel-median Intent per condition is unchanged, and the ToM mean Intent drifts slightly downward under anchor blinding (consistent with the informed judge being mildly anchored upward by the disclosed Castelli mean). Per-tester strict-rank changes are minor reshuffles at the gemini-3.5-flash boundary. The under-attribution and ToM-toward-HF-ASD findings hold under both variants; the bias was conservative.}
\label{tab:judge_blinding}
\end{table*}

\paragraph{Canonical version reported in the main text.}
The main-text animated-triangles numbers in Sections~\ref{sec:benchmarkB_results} and \ref{sec:cross_task}, the per-model profiles in Figure~\ref{fig:benchmarkB_tom_profile}, the Y axis of Figure~\ref{fig:joint_dissociation}, the cross-tabulation in Table~\ref{tab:nearest_ref_a2}, and the per-condition distances in Tables~\ref{tab:benchmarkB_distances} and \ref{tab:per_condition_distances} are all reported under the anchor-blinded L1 canonical rubric.
The informed-judge numbers are retained in Table~\ref{tab:judge_blinding} above as the audit-trail comparison and are not used elsewhere in the paper.
Direct human ratings on the specific $9$-tester $\times$ $12$-clip $\times$ $3$-judge production cells were not collected and remain a useful next step for further tightening the LLM-as-rater versus human-rater asymmetry.

\paragraph{Same-family judge audit.}
The cost-efficient three-judge production panel shares model families with three of the nine testers (claude-opus-4.7 and claude-sonnet-4.6 vs the claude-haiku-4.5 judge, gemini-3.0-pro and gemini-3.5-flash vs the gemini-2.5-flash judge, and qwen-3.5-plus vs the qwen2.5-72b-instruct judge), yielding 60 of 324 in-family (judge, tester) cells.
Excluding these 60 in-family cells from the per-condition median Intent aggregation under the anchor-blinded canonical rubric on the numbered-grid stimulus leaves the ToM, GD, and Random condition medians exactly unchanged ($\Delta = 0.0$ throughout).
Same-family judge bias is therefore empirically null at the panel-median aggregation level on this data.

\section{Stimulus-Format Ablation (Plain Grid vs Numbered Grid)}
\label{sec:appendix_ablation}

\paragraph{Rationale.}
The canonical $4 \times 4$ frame grid asks the model to infer that the sixteen cells are temporally ordered samples of a single short clip.
Models with weaker fine-grained visual reasoning may be unable to recover this temporal order from the static composite alone, and that vision-side failure could confound any inference about intent-attribution capacity.
We therefore evaluate a second stimulus format in which each of the sixteen cells carries an explicit $1{-}16$ index badge in its corner, giving the model the temporal order as a free signal.
Everything else, including the prompt text byte-for-byte, the rubric, the rater jury, the panel of nine testers, the twelve canonical Frith-Happ\'e clips, the sixteen-frame sampling, and $K_\text{tester} = 3$, is held identical.
The ablation isolates whether observed Intent under-attribution reflects temporal-parsing failure on the vision side or intent-attribution failure on the ToM side.

\paragraph{Headline comparison.}
Table~\ref{tab:ablation_headline} reports the v1 (plain grid) versus v2 (numbered grid) headline metrics on the same $9 \times 12 \times 3 = 324$ (tester, clip, judge) cells per arm, each aggregating over $K_\text{tester} = 3$ trials.

\begin{table*}[t]
\centering
\small
\begin{tabular}{lccc}
\toprule
Metric & v1 plain & v2 numbered & $\Delta$ \\
\midrule
Strict rank $\text{ToM} > \text{GD} > \text{Random}$ testers & 0 / 9 & 5 / 9 & $+5$ \\
Panel median Intent on ToM & $2.0$ & $2.0$ & $0.0$ \\
Panel median Intent on GD & $2.0$ & $2.0$ & $0.0$ \\
Panel median Intent on Random & $1.0$ & $1.0$ & $0.0$ \\
Panel mean (Intent, Approp) on ToM & $(2.25, 0.95)$ & $(2.31, 0.94)$ & --- \\
Distance from panel mean on ToM to TD-adult & $2.18$ & $2.13$ & $-0.05$ \\
Distance from panel mean on ToM to HF-ASD-adult & $0.79$ & $0.73$ & $-0.06$ \\
\bottomrule
\end{tabular}
\caption{v1 (plain $4 \times 4$ grid) versus v2 (numbered $4 \times 4$ grid) on the same nine testers, twelve clips, and three judges with $K_\text{tester} = 3$, anchor-blinded canonical rubric per Appendix~\ref{sec:appendix_judging}. The numbered-grid stimulus moves five testers into strict Intent rank, but the panel median Intent and the Stage-C nearest-group verdict per condition are unchanged.}
\label{tab:ablation_headline}
\end{table*}

\paragraph{Per-tester strict-rank flips.}

\begin{table*}[t]
\centering
\small
\begin{tabular}{lcccc}
\toprule
Tester & v1 ToM / GD / Rand & v1 rank & v2 ToM / GD / Rand & v2 rank \\
\midrule
claude-opus-4.7        & $3.0 / 3.0 / 2.0$ & no  & $3.0 / 2.0 / 1.5$ & yes  \\
claude-sonnet-4.6      & $1.5 / 2.0 / 0.0$ & no  & $1.0 / 2.0 / 1.0$ & no   \\
gemini-3.0-pro         & $3.0 / 3.0 / 2.0$ & no  & $3.0 / 2.0 / 1.0$ & yes  \\
gemini-3.5-flash       & $3.0 / 2.0 / 2.0$ & no  & $3.5 / 2.5 / 2.0$ & yes  \\
gpt-5.4                & $2.0 / 2.0 / 0.5$ & no  & $2.0 / 2.0 / 1.0$ & no   \\
gpt-5.5                & $2.0 / 2.0 / 1.0$ & no  & $3.0 / 2.0 / 1.0$ & yes  \\
grok-4.3               & $1.0 / 1.0 / 0.0$ & no  & $1.0 / 1.0 / 0.0$ & no   \\
kimi-k2.6              & $2.0 / 2.0 / 1.0$ & no  & $2.0 / 2.0 / 1.0$ & no   \\
qwen-3.5-plus          & $2.0 / 2.0 / 0.0$ & no  & $3.0 / 2.0 / 0.5$ & yes  \\
\bottomrule
\end{tabular}
\caption{Per-tester strict Intent rank $\text{ToM} > \text{GD} > \text{Random}$ status under plain (v1) and numbered (v2) grid, anchor-blinded canonical rubric. No tester reaches strict rank under v1; five testers (opus, gemini-3.0-pro, gemini-3.5-flash, gpt-5.5, qwen) reach strict rank under v2. The five gainers cluster as the high-vision frontier subset of the panel.}
\label{tab:ablation_flips}
\end{table*}

\paragraph{Stage-C nearest-group verdict robustness.}
Under both v1 and v2 the panel mean (Intent, Approp) profile on the ToM condition is closer to the HF-ASD-adult group mean than to the TD-adult group mean (v1 distances $0.79$ versus $2.18$; v2 distances $0.73$ versus $2.13$), and on both GD and Random the panel mean is closer to TD-adult.
The per-condition nearest-group verdict is therefore identical across the two stimulus formats; ToM nearest HF-ASD-adult, GD and Random nearest TD-adult.
Our central descriptive finding does not depend on the stimulus-format choice.

\paragraph{Why v2 is canonical for the main text.}
The numbered-grid (v2) is canonical because removing the temporal-parsing confound is a pre-defined design requirement; any animated-triangles benchmark used to claim a ToM-side limitation must first rule out that the observed Intent under-attribution is driven by failure to recover frame order from a static composite.
The plain grid (v1) does not rule this out, so we retain it as the ablation arm that quantifies how much of the panel-level Intent collapse is recoverable under explicit temporal cueing.
The five-of-nine strict-rank recovery under v2 is the empirical confirmation that v2 dissolves the vision-side confound, not the reason we chose v2.

\section{Frame Representation Selection}
\label{sec:appendix_frame_selection}

\paragraph{Motivation.}
The animated-triangles benchmark presents each clip as a single composite image of 16 uniformly sampled frames arranged in a $4 \times 4$ grid.
This representation was chosen against eight alternatives (sequence of $N$ \texttt{image\_url} payloads, larger grids, native video) to maximize agreement with the model's full-fidelity video path while keeping per-cell prompt cost tractable across the $9 \times 12 \times 3$ judge cells of the main run.

\paragraph{Reference baseline.}
Of the nine frontier VLM testers in the main panel, only Gemini 2.5 Pro accepts a native \texttt{video\_url} payload.
We treat Gemini's native-video path as the 100\% reference.
It routes the full mp4 through Gemini's video tokens, populates the \texttt{usage.\allowbreak prompt\_tokens\_\allowbreak details.\allowbreak video\_tokens} field non-trivially, and is the same path the model was trained for.

\paragraph{Conditions tested.}
On the same Gemini 2.5 Pro tester and three Castelli clips (one Random, one GD, one ToM), nine frame representations were evaluated: \texttt{native\_video} (reference), three sequence variants (\texttt{frames\_16}, \texttt{frames\_32}, \texttt{frames\_64}) and five grid variants (\texttt{grid\_4x4} / \texttt{4x8} / \texttt{8x8} at fixed $320 \times 240$ per-cell resolution, plus \texttt{grid\_4x4\_fullres} and \texttt{grid\_8x8\_fullres} at source resolution).
The same three flagship LLM judges (claude-opus-4.7, gpt-5, gemini-2.5-pro) at $K_\text{judge}=3$ scoring reps applied the Castelli rubric.
The top three candidates (\texttt{frames\_64}, \texttt{grid\_4x4}, \texttt{native\_video}) were promoted to $K_\text{tester}=3$ to tighten the verdict beyond $K=1$ noise.

\paragraph{Result.}
We define per-condition similarity to \texttt{native\_video} as the equal-weighted average of Intent and Approp similarity percentages, each derived from per-clip mean absolute error against \texttt{native\_video} on the rubric's full scale.
Table~\ref{tab:frame_selection} reports the final ranking.
\texttt{frames\_64} achieves the highest similarity to native at 88.2\%, but at 16{,}594 prompt tokens per cell.
\texttt{grid\_4x4} achieves 85.8\% similarity at 1{,}416 prompt tokens per cell, an $11.7\times$ reduction in token cost for a 2.4 pp reduction in similarity.
We adopt \texttt{grid\_4x4} as the canonical frame representation.
It is Pareto-dominant on similarity-per-token over both the higher-fidelity sequence representations and the larger grids.

\begin{table*}[t]
\centering
\small
\begin{tabular}{llcccc}
\toprule
Condition & Family & Intent sim & Approp sim & Total sim & Tokens/cell \\
\midrule
\texttt{native\_video} (reference)   & video           & 100.0\% & 100.0\% & 100.0\%        & 11571 \\
\midrule
\texttt{frames\_64}                  & sequence        & 89.7\%  & 86.8\%  & 88.2\%         & 16594 \\
\textbf{\texttt{grid\_4x4} (chosen)} & grid 320$\times$240 & 85.2\% & 86.4\%  & \textbf{85.8\%} & \textbf{1416} \\
\texttt{frames\_16}                  & sequence        & 71.7\%  & 87.9\%  & 79.8\%         & 4188 \\
\texttt{grid\_8x8\_fullres}          & grid native-res & 64.5\%  & 84.3\%  & 74.4\%         & 6210 \\
\texttt{frames\_32}                  & sequence        & 67.5\%  & 69.5\%  & 68.5\%         & 8377 \\
\texttt{grid\_4x8}                   & grid 320$\times$240 & 67.7\% & 47.8\%  & 57.8\%         & 2820 \\
\texttt{grid\_4x4\_fullres}          & grid native-res & 54.1\%  & 42.3\%  & 48.2\%         & 1545 \\
\texttt{grid\_8x8}                   & grid 320$\times$240 & 61.0\% & 20.9\%  & 41.0\%         & 5648 \\
\bottomrule
\end{tabular}
\caption{Frame representation similarity to Gemini's native video path. Similarity is the equal-weighted mean of per-clip Intent and Approp similarity percentages, each derived from mean absolute error against \texttt{native\_video} on the rubric's full scale (Intent 0--5, Approp 0--3). We adopt \texttt{grid\_4x4} as canonical: 85.8\% similarity at 1{,}416 tokens per cell, $11.7\times$ cheaper than \texttt{frames\_64} (88.2\% at 16{,}594 tokens). The top three candidates were run at $K_\text{tester}=3$ to tighten the verdict beyond $K=1$ noise; the remaining five used $K_\text{tester}=1$.}
\label{tab:frame_selection}
\end{table*}

\section{Rubric and Judge Validation Battery (Phase 1)}
\label{sec:appendix_rubric_validation}

\paragraph{Anchor set.}
Before the main 9-tester $\times$ 12-clip production run, we validate that the \citet{castelli_2000} Appendix-2 rubric is interpretable by LLM judges on human-anchored exemplars.
The anchor set consists of 14 paper-derived ground-truth items, namely five ToM-condition transcripts T1--T5 lifted verbatim from \citet{castelli_2002} (human participants' free-text descriptions of the canonical clips, with expected Intent in 0--5 and Approp in 0--3) and nine Appendix-2 exemplar phrases A0a--A5b from \citet{castelli_2000} covering the Random / GD / interaction range with expected Intent or Approp targets per Castelli's worked examples.

\paragraph{Battery design.}
Three flagship LLM judges (claude-opus-4.7, gpt-5, gemini-2.5-pro) score each anchor under the Castelli rubric template (SHA pinned, \texttt{llm\_judge\_prompt.yaml}) at $K=5$ repetitions, $T=0$, no CoT.
Total: $14 \times 3 \times 5 = 210$ calls.
The double gate is (i) per-judge recovery percentage $\geq 80\%$ (a cell counts as recovered iff predicted score lies within $\pm 0.5$ of the expected interval), and (ii) per-anchor standard deviation $\leq 0.5$ on Intent and $\leq 0.8$ on Approp across the 5 reps.
The SD gate measures within-judge stochasticity not eliminated by $T=0$.

\paragraph{Result.}
All three flagship judges PASS both gates.
Intent recovery is 100\% (claude-opus-4.7), 100\% (gpt-5), 100\% (gemini-2.5-pro); Approp recovery is 100\%, 96\%, 100\% respectively.
Maximum per-anchor SDs are 0.40 / 0.49 (opus), 0.00 / 0.40 (gpt-5), 0.00 / 0.43 (gemini-pro) for Intent / Approp; all within gate.
The validated rubric and prompt SHA are then frozen and copied into the production rater; the production aggregation pipeline asserts SHA match at startup, blocking silent drift.

\section{Judge Model Ablation}
\label{sec:appendix_judge_ablation}

\paragraph{Motivation.}
At production scale ($9 \times 12 \times 3 \times K_\text{tester}=3$) flagship inference dominates the per-experiment budget; we therefore ablate whether a cheaper cross-vendor judge panel can match flagship recovery on the same 14-anchor battery used in Appendix~\ref{sec:appendix_rubric_validation}.

\paragraph{Candidates.}
Seven cheaper candidates spanning five vendor families (Anthropic, OpenAI, Google, Moonshot, DeepSeek, Qwen) were evaluated against the same 14-anchor protocol, namely claude-sonnet-4.6, claude-haiku-4.5, gpt-5-mini, gemini-2.5-flash, kimi-k2, deepseek-v3, qwen2.5-72b-instruct.
Total: $7 \times 14 \times 5 = 490$ calls at $T=0$.
The same double gate from Phase 1 applies.

\paragraph{Result.}
Six of seven candidates PASS both gates.
The single failure is gpt-5-mini (Approp recovery 80\%, max Approp SD 1.47); the failure mode is a magnified form of the same instability the flagship gpt-5 already exhibits on long ambiguous inputs.
Three candidates --- claude-haiku-4.5, gemini-2.5-flash, qwen2.5-72b-instruct --- achieve 100\% recovery with per-anchor Intent and Approp SDs equal to 0 across all 14 anchors $\times$ 5 reps (Table~\ref{tab:judge_ablation}).
These three are strictly more stable than any of the three flagship judges.
We adopt this trio as the production scoring panel.
The trio spans three distinct training lineages (Anthropic instruction-following, Google reasoning, Alibaba's open-source Qwen), reducing family-correlated reading bias; it costs \$0.0129 per cell versus \$0.0676 per cell on the flagship panel (81\% reduction); and it shows SD=0 ceilings on every anchor.

\begin{table*}[!t]
\centering
\footnotesize
\begin{tabular}{llcccccc}
\toprule
Family & Judge & Intent \% & Approp \% & Max I-SD & Max A-SD & Verdict & \$/call \\
\midrule
Anthropic & claude-opus-4.7 (flagship)            & 100  & 100 & 0.40          & 0.49          & PASS & 0.0400 \\
Anthropic & claude-sonnet-4.6                     & 100  & 100 & \textbf{0.00} & \textbf{0.00} & PASS & 0.0240 \\
Anthropic & \textbf{claude-haiku-4.5 (chosen)}    & 100  & 100 & \textbf{0.00} & \textbf{0.00} & PASS & 0.0080 \\
OpenAI    & gpt-5 (flagship)                      & 100  & 96  & 0.00          & 0.40          & PASS & 0.0138 \\
OpenAI    & gpt-5-mini                            & 95.7 & 80  & 0.49          & 1.47          & FAIL & 0.0028 \\
Google    & gemini-2.5-pro (flagship)             & 100  & 100 & 0.00          & 0.43          & PASS & 0.0138 \\
Google    & \textbf{gemini-2.5-flash (chosen)}    & 100  & 100 & \textbf{0.00} & \textbf{0.00} & PASS & 0.0034 \\
Moonshot  & kimi-k2                               & 100  & 100 & 0.40          & 0.00          & PASS & 0.0043 \\
DeepSeek  & deepseek-v3                           & 100  & 100 & 0.40          & 0.40          & PASS & 0.0014 \\
Qwen      & \textbf{qwen2.5-72b-instruct (chosen)}& 100  & 100 & \textbf{0.00} & \textbf{0.00} & PASS & 0.0015 \\
\bottomrule
\end{tabular}
\caption{Judge model ablation against the 14-anchor Phase-1 battery. Three lower-cost candidates (claude-haiku-4.5, gemini-2.5-flash, qwen2.5-72b-instruct) achieve 100\% recovery with per-anchor SD = 0, matching flagship within-judge consistency at $T=0$ at lower per-call cost; we adopt the trio as the production scoring panel for the main 9-tester $\times$ 12-clip $\times$ 3-judge run. Pricing is OpenRouter list price (3000 input + 1000 output tokens per call). gpt-5-mini is the only candidate to FAIL; its Approp SD of 1.47 across 5 reps at $T=0$ amplifies the same instability the flagship gpt-5 shows on long ambiguous inputs.}
\label{tab:judge_ablation}
\end{table*}

\section{Reproducibility Details}
\label{sec:appendix_reproducibility}
Released with this submission are the benchmark stimuli, prompts, the SHA-locked scoring rubric, judge configuration, the two-firewall harness, the aggregation scripts that produce every figure and table in this paper, and a digest manifest of the runs underlying the reported numbers.
Released with the camera-ready (held back at submission for review-time blinding and storage-quota reasons) are the full per-(tester, clip, judge) raw scoring outputs and the full per-group source list for the human-reference fallback rule in Appendix~\ref{sec:appendix_anchor_sensitivity}.

\end{document}